\RequirePackage{luatex85}

\documentclass{article}
\usepackage{ijcai18}

\usepackage{times}
\usepackage{xcolor}
\usepackage{soul}
\usepackage[utf8]{inputenc}
\usepackage[small]{caption}
\usepackage{graphicx}
\graphicspath{ {figs/} }
\usepackage{amsmath}
\usepackage{amssymb}
\usepackage{algorithm}
\usepackage[noend]{algpseudocode}
\usepackage{setspace}
\usepackage{hyperref}
\usepackage{float}
\usepackage{tabularx}
\usepackage{booktabs}
\usepackage{dblfloatfix}
\usepackage{siunitx}
\usepackage{todonotes}
\usepackage[normalem]{ulem}
\usepackage{cleveref}
\usepackage{verbatim}

\hypersetup{
    colorlinks = true,
    citecolor = blue!50!black,
}

\usepackage{amsmath,amsfonts,bm}

\def\eqref#1{equation~\ref{#1}}

\def\1{\bm{1}}

\def\vh{{\bm{h}}}

\def\vz{{\bm{z}}}

\DeclareMathAlphabet{\mathsfit}{\encodingdefault}{\sfdefault}{m}{sl}
\SetMathAlphabet{\mathsfit}{bold}{\encodingdefault}{\sfdefault}{bx}{n}

\newcommand{\R}{\mathbb{R}}

\renewcommand{\vz}{\mathbf{z}}
\renewcommand{\vh}{\mathbf{h}}
\renewcommand{\b}{\mathbf}

\renewcommand{\R}{\mathbb{R}}

\title{Incorporating Literals into Knowledge Graph Embeddings} 

\author{
Agustinus Kristiadi\thanks{Equal contribution}$^4$,
Mohammad Asif Khan$^*$$^1$,
Denis Lukovnikov$^1$, 
Jens Lehmann$^{1,2}$,
Asja Fischer$^3$
\\
$^1$ SDA Group, University of Bonn \\
$^2$ EIS Department, Fraunhofer IAIS \\
$^3$ Ruhr-University Bochum \\
$^4$ University of T\"{u}bingen \\
agustinus.kristiadi@uni-tuebingen.de,
s6mokhan@uni-bonn.de,
lukovnik@cs.uni-bonn.de,\\
jens.lehmann@uni-bonn.de,
jens.lehmann@iais.fraunhofer.de,
asja.fischer@rub.de
}

\begin{document}

\maketitle

\begin{abstract}
Knowledge graphs are composed of different elements: entity nodes, relation edges, and literal nodes. Each literal node contains an entity's attribute value (e.g.~the \textit{height} of an entity of type \textit{person}) and thereby encodes information which in general cannot be represented by relations between entities alone. However, most of the existing embedding- or latent-feature-based methods for knowledge graph analysis only consider entity nodes and relation edges, and thus do not take the information provided by literals into account. In this paper, we extend existing latent feature methods for link prediction by a simple portable module for incorporating literals, which we name LiteralE. Unlike in concurrent methods where literals are 
incorporated by adding a literal-dependent term to the output of the scoring function and thus  only indirectly affect the entity embeddings, LiteralE directly enriches these embeddings with information from literals via a learnable parametrized function. 
This function can be easily integrated into the scoring function of existing methods 
and learned along with the entity embeddings in an end-to-end manner. In an extensive empirical study over three datasets, we evaluate LiteralE{-extended} versions of various state-of-the-art latent feature methods for link prediction and demonstrate that 
LiteralE presents an effective way to improve their performance.
For these experiments, we augmented standard datasets with their literals, which we publicly provide as testbeds for further research.
Moreover, we show that LiteralE leads to an qualitative improvement of the embeddings
and that it can be easily extended to handle literals from different modalities. 
\end{abstract}

\section{Introduction}

\begin{figure}[t]
  \centering
  \includegraphics[width=0.7\columnwidth]{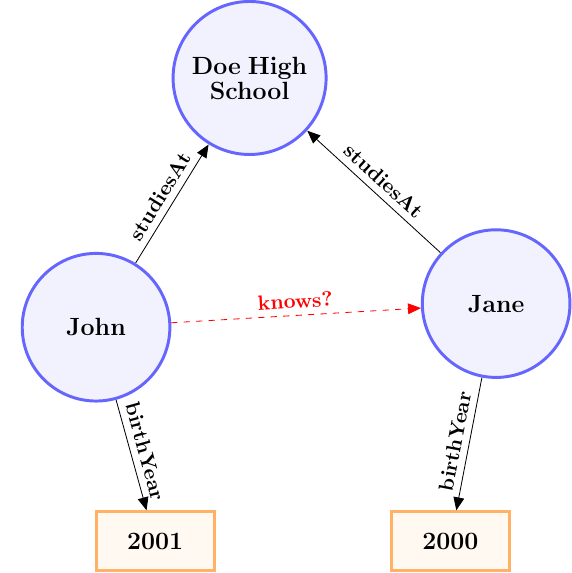}
  \caption{\label{fig:kg_lit} Literals (box) encode information that cannot be represented by relations alone, and are useful for link prediction task. For instance, by considering both \texttt{birthYear} literals and the fact that \texttt{John} and \texttt{Jane} both study at \texttt{Doe High School}, we can be more confident that the relation \texttt{knows} between \texttt{John} and \texttt{Jane} exists.}
\end{figure}

Knowledge graphs (KGs) form the backbone of a range of applications, for instance in the areas of search, question answering and data integration. Some well known KGs are DBpedia~\cite{lehmann2015dbpedia}, Freebase~\cite{bollacker2008freebase}, YAGO3~\cite{mahdisoltani2014yago3}, and the Google Knowledge Graph~\cite{dong2014knowledge}.
There are different knowledge representation paradigms for modeling KGs such as the Resource Description Framework (RDF) and (labeled) property graphs. 
Within this paper, we consider a KG to be a set of triples, where each triple connects an entity (shown as circle in \Cref{fig:kg_lit}) to another entity or a literal (the latter shown as rectangle in \Cref{fig:kg_lit}) via relationships. 
Such KGs can be represented by the RDF and property graph paradigms, i.e.~the methods presented in this paper are applicable to both. 
To give a 
concrete example, the KG depicted in \Cref{fig:kg_lit} includes the triples (\texttt{John}, \texttt{Doe High School}, \texttt{studiesAt}) and (\texttt{Jane}, \texttt{2000}, \texttt{birthYear}). The first triple expresses the relationship between an entity and another entity. The second triple expresses a relationship between an entity and a literal\footnote{For more information about the RDF concepts see \url{https://www.w3.org/TR/rdf11-concepts}
}.

Knowledge graphs aim to capture factual knowledge within a particular domain. 
However, they are often incomplete since, e.g., more information is provided for popular than for unknown  entities or because the KG is partially or fully generated via an automatic extraction process. 
As a result, KGs rely heavily on methods predicting unknown triples given all known triples. 
This problem is usually referred to as link prediction. 
The closely related problem of detecting incorrect triples in KGs is referred to as link correction and is relevant for improving the quality of a KG.  

Due to the importance of the problem, many methods for link prediction and correction in KGs have been developed. 
The two main classes of these methods are graph feature and latent feature methods~\cite{nickel2016review}.
Graph feature methods predict the existence of triples based on 
features directly observed in the KG, such as the neighborhood of an entity and paths to other entities. 
They are well suited for modeling local graph patterns. 
In latent feature models, low-dimensional, latent representations (also called embeddings) of entities and relations are learned. 
These embeddings incorporate the KG structure, can capture global patterns, and allow to compute the likeliness of a given triple in terms of a probability or score function. 
However, most of the recent 
work on latent feature models only takes entities and their relations to other entities into account. 
Therefore, they are missing the additional information encoded in literals.  
For example, \Cref{fig:kg_lit} shows two entities with both structural (visiting the same school) as well as literal (birth years) information. 
To maximize the accuracy of predicting a \texttt{knows} relation between these entities, structural and literal information should be combined as people visiting the same school and 
having
similar age tend to have a higher probability of knowing each other.

In this paper, we investigate the advantage obtained by incorporating the additional information provided by literals into latent feature methods.
We introduce LiteralE, a method to enrich entity embeddings with their literal information. 
Given an entity embedding, we incorporate its corresponding literals using a learnable parametric function,
which gets the vanilla embedding and the entity's literals as input, and outputs a literal-enriched embedding.
This embedding can then replace the vanilla embedding in any latent feature model, without changing its original scoring function and the resulting system can be jointly trained with stochastic gradient descent, or any other gradient based algorithm of choice, in an end-to-end manner. 
 Therefore, LiteralE can be seen as an extension module that can be universally combined with 
 any existing latent feature method.
Within this paper, we mainly focus on numerical literals.
However, we demonstrate that the principle can be directly generalized to other literal types, such as textual and image information, e.g.~by providing low-dimensional vector representation of image or text~\cite{xie2016representation,xu2016knowledge} as an additional input to LiteralE. 

Our contributions in this paper are threefold:
\begin{itemize}
 \item We introduce LiteralE, a universal approach to enrich latent feature methods
 with literal information via a learnable parametric function. In contrast to other latent feature models including literals, our approach does not require specific prior knowledge, does not rely on a fixed function to combine entity embeddings and literals, can model interactions between an embedding of an entity and all its literal values and can be trained end-to-end. 
\item We evaluate LiteralE on standard link prediction datasets:  \textsf{FB15k}, \textsf{FB15k-237} and \textsf{YAGO3-10}. We extended  \textsf{FB15k} and \textsf{FB15k-237} with literals, in order to allow for direct comparison against other methods on these standard datasets. We provide these literal-extended versions (augmented with numerical and textual literals) 
and hope they can serve as a testbed for future research on the inclusion of literals in KG modeling.\footnote{A literal-extended version of \textsf{YAGO3-10} is provided by~\cite{pezeshkpour2017embedding}.}
\item Based on experimental results on the extended datasets, we show that exploiting the information provided by literals significantly increases the link prediction 
performance of existing latent feature  methods 
as well as the quality of their embeddings.
\end{itemize}

This paper is organized as follows. In \Cref{sec:prelim} we review several latent feature methods for link prediction in KGs. In \Cref{sec:literale} we present LiteralE, our approach for incorporating literals into existing latent feature methods. We
give a brief review of the related literatures and
contrast LiteralE with other methods incorporating literals 
in \Cref{sec:related}. Our experiment methodology is described in \Cref{sec:experiment}, and in \Cref{sec:results} we present our experiment results. Finally, we conclude our paper in \Cref{sec:conclusion}. 

Our implementation of the proposed methods and all datasets are publicly available at: \url{https://github.com/SmartDataAnalytics/LiteralE}.

\section{Preliminaries}
\label{sec:prelim}

In the following we will describe the link prediction problem more formally and give a brief overview over well-known latent feature methods. 

\subsection{Problem Description}
Link prediction is defined as the task of deciding whether a fact (represented by a triple) is true or false given a KG.
More formally, let $\mathcal{E} = \{ e_1, \cdots, e_{N_e} \}$ be the set of entities, $\mathcal{R} = \{r_1, \cdots, r_{N_r} \}$ be the set of relations connecting two entities, $\mathcal{D}= \{d_1, \cdots, d_{N_d} \}$ be the set of relations connecting an entity and a literal, i.e., the data relations, and $\mathcal{L}$ be the set of all literal values.
A knowledge graph $\mathcal{G}$ is a subset of 
$(\mathcal{E} \times \mathcal{E} \times \mathcal{R}) \cup (\mathcal{E} \times \mathcal{L} \times \mathcal{D})$ representing the facts that are assumed to hold.
Link prediction can be formulated by a function $\psi: \mathcal{E} \times \mathcal{E} \times \mathcal{R} \to \mathbb{R}$
mapping 
each possible fact represented by the corresponding triple $(e_i, e_j, r_k) \in$ 
$ \mathcal{E} \times \mathcal{E} \times \mathcal{R}$ to a score value, where a higher value
implies the triple is more likely to be true.

\subsection{Latent Feature Methods}

In general, latent feature methods are a class of methods in which low dimensional vector representations of entities and relations, called \textit{embeddings} or \textit{latent features}, are learned.
Let $H$ be the embedding dimension.
We define a score function $f: \mathbb{R}^H \times \mathbb{R}^H \times \mathbb{R}^H \to \mathbb{R}$ 
that maps a triple of embeddings $(\mathbf{e}_i, \mathbf{e}_j, \mathbf{r}_k)$
to a score $f(\mathbf{e}_i, \mathbf{e}_j, \mathbf{r}_k)$ that correlates with the truth value of the triple.
In latent feature methods, the score of any triple $(e_i, e_j, r_k) \in \mathcal{E} \times \mathcal{E} \times \mathcal{R}$ is then defined as $\psi(e_i, e_j, r_k) \overset{\underset{\mathrm{def}}{}}{=} f(\mathbf{e}_i, \mathbf{e}_j, \mathbf{r}_k)$. 

Latent feature methods for link predictions are well studied. 
These methods follow a score-based approach as described above but make use of different kind of scoring functions $f$. 
In this paper we study the 
potential benefit
of incorporating numerical literals in three state of the art methods: DistMult~\cite{dong2014knowledge}, ComplEx~\cite{trouillon2016complex}, and ConvE~\cite{dettmers2018conve}, which are described in the following.
Note however, that these are just an exemplary choice of methods and our approach for incorporating literals
can easily be adopted to 
other latent feature methods.

The \textbf{DistMult} scoring function is defined as diagonal bilinear interaction 
between the two entity embeddings and the relation embedding corresponding to a given triple, as follows
\begin{equation} \label{eq:distmult}
f_{\text{DistMult}}(\mathbf{e}_i, \mathbf{e}_j, \mathbf{r}_k) = \langle \mathbf{e}_i, \mathbf{e}_j, \mathbf{r}_k \rangle = 
\mathbf{e}_i^\intercal \, \text{diag}(\mathbf{r_k}) \, \mathbf{e}_j \enspace .
\end{equation}
%
Observe that 
DistMult is cheap to implement, both in terms of computational and space complexity.


\textbf{ComplEx} can be seen as DistMult analogue in the complex space. The embedding vectors have two parts: the real part $\text{Re}(\mathbf{e})$ and $\text{Re}(\mathbf{r})$, and the imaginary part $\text{Im}(\mathbf{e})$ and $\text{Im}(\mathbf{r})$, respectively. The scoring function is defined as
\begin{equation}
  \label{eq:complex}
  \begin{aligned}
  f_{\text{ComplEx}}(\mathbf{e}_i, \mathbf{e}_j, \mathbf{r}_k) &= \text{Re}({\langle \mathbf{e}_i,\mathbf{\bar{e}}_j,\mathbf{r}_k\rangle})\\
  &=\langle \text{Re}(\mathbf{e}_i), \text{Re}(\mathbf{e}_j), \text{Re}(\mathbf{r}_k) \rangle \\
  &+ \langle \text{Im}(\mathbf{e}_i), \text{Im}(\mathbf{e}_j), \text{Re}(\mathbf{r}_k) \rangle \\
  &+ \langle \text{Re}(\mathbf{e}_i), \text{Im}(\mathbf{e}_j), \text{Im}(\mathbf{r}_k) \rangle \\
  &- \langle \text{Im}(\mathbf{e}_i), \text{Re}(\mathbf{e}_j), \text{Im}(\mathbf{r}_k) \rangle \enspace .
  \end{aligned}
\end{equation}
ComplEx thus has twice the number of parameters compared to DistMult but provides the benefit of modeling asymmetric relationships better, as discussed by~\cite{trouillon2016complex}.

\textbf{ConvE} 
employs a convolutional neural network
to extract features from entity and relation embeddings. Let $h$ be a nonlinear function, $\omega\in\mathcal{R}^{k\times m\times n}$ be convolution filters, and $\mathbf{W}\in\mathcal{R}^{kmn\times H}$ be a weight matrix. The ConvE score function is then defined by
\begin{equation}
  \label{eq:conve}
  f_{\text{ConvE}}(\mathbf{e}_i, \mathbf{e}_j, \mathbf{r}_k) = h(\textbf{vec}(h([\mathbf{e}_i, \mathbf{r}_k] \ast \omega)) \mathbf{W})  \, \mathbf{e}_j \enspace,
\end{equation}
where \textbf{vec}$(\cdot)$ is the vectorization of output of convolutional filters. By employing deep feature extractors in the form of nonlinear convolutional layers, ConvE is able to encode more expressive features while remaining highly parameter efficient.

\section{LiteralE}
\label{sec:literale}

Our method of incorporating literals into existing latent feature methods, which we call LiteralE, is a simple, modular, and universal extension which can potentially enhance the performance of 
arbitrary latent feature methods.

Let $\mathbf{L} \in \mathbb{R}^{N_e \times N_d}$ be a matrix, where each entry $\mathbf{L}_{ik}$ contains the $k$-th literal value of the $i$-th entity if a triple with the $i$-th entity and the $k$-th data relation exists in the KGs,
and zero otherwise. 
We will refer to the $i$-th row $\mathbf{l}_i$ of $\mathbf{L}$ as the literal vector of the $i$-th entity.
As an illustration, consider the KG part depicted in~\Cref{fig:kg_lit} and imagine that there only exist three data relations in this specific KG: \texttt{heightCm}, \texttt{birthYear}, and \texttt{countryArea}. For the 
entity \texttt{John} 
we will then have the literal vector $(0, 2001, 0)$ in the particular row corresponding to \texttt{John} in matrix $\mathbf{L}$, as \texttt{John} only has literal information for \texttt{birthYear}.\footnote{Note that in practice, we normalize the literal values.}

At the core of LiteralE is a function 
$g: \mathbb{R}^H \times \mathbb{R}^{N_d} \to \mathbb{R}^H$
that takes an entity's embedding and a literal vector as inputs and maps them to 
a vector of the same dimension as the entity embedding.
%
This vector forms an literal-enriched embedding vector
that can replace the original embedding vector in the scoring function of any latent feature model. 
For example, in our experiments, we replace every entity embedding $\mathbf{e}_i$ with $\mathbf{e}_i^\text{lit} = g(\mathbf{e}_i, \mathbf{l}_i)$ in the scoring functions of DistMult and ConvE.
For ComplEx, where the embeddings have a real and an imaginary part, 
we use two separate functions to map $\text{Re}(\mathbf{e}_i)$ and $\text{Im}(\mathbf{e}_i)$ to their literal-extended counterparts.
Aside of these changes regarding the entity embeddings, the score functions are the same as described before in \cref{eq:distmult},~\cref{eq:complex}, and~\cref{eq:conve}. For instance, the LiteralE-extended version of DistMult is given by $f_{\text{DistMult}}(\mathbf{e}^{\text{lit}}_i, \mathbf{e}^{\text{lit}}_j, \mathbf{r}_k)$.

\begin{figure}[t]
  \centering
  \includegraphics[width=0.8\columnwidth]{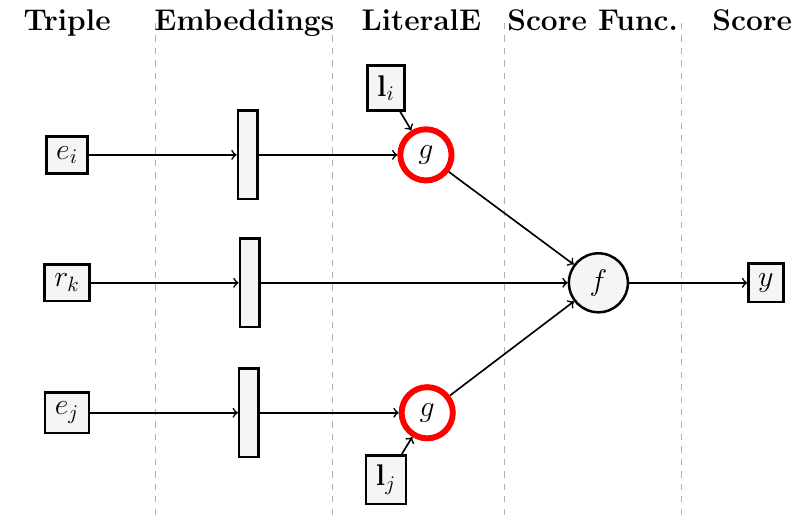}
  \caption{\label{fig:literalE} Overview on how LiteralE is applied to the base scoring function $f$. LiteralE takes the embedding and the corresponding literals as input, and combines them via a learnable function $g$. The output is a joint embedding which is further used in the score function $f$.}
\end{figure}

We will now describe the function $g$ in detail. First, since we would like $g$ to be flexible, we need it to be learnable. Second, we would like $g$ to be able to decide whether the additional literal information is useful or not, and adapt accordingly, e.g. by incorporating or ignoring that information. We therefore take cue from the gating mechanism present in RNNs, such as the gated recurrent unit (GRU)~\cite{cho2014gru}, and let $g$ be defined by
\begin{align}
    \label{eq:g_def}
    &g: \mathbb{R}^H \times \mathbb{R}^{N_d} \to \mathbb{R}^H \nonumber \\
        &\b{e}, \b{l} \mapsto \vz \odot \vh + (1-\vz) \odot \b{e} \, ,
\end{align}
where $\odot$ is the pointwise multiplication and 
\begin{align}
    \label{eq:g_detail}
    \vz &= \sigma(\b{W}_{ze}^\text{T} \b{e} + \b{W}_{zl}^\text{T} \b{l} + \b{b}) \nonumber \\
    \vh &= h(\b{W}_h^\text{T} [\b{e}, \b{l}]) \, .
\end{align}
Note that $\b{W}_h \in \R^{H+N_d \times H}$, $\b{W}_{ze} \in \R^{H \times H}$, $\b{W}_{zl} \in \R^{N_d \times H}$, and $\b{b} \in \R^{H}$ are the parameters of $g$, $\sigma$ is the sigmoid function, and $h$ is a component-wise nonlinearity (e.g. the hyperbolic tangent). 

LiteralE introduces some overhead in the number of parameters compared to the base method. This overhead is equal to the number of parameters of the function $g$ 
and is compared to that of other approaches for the incorporation of literals in \Cref{tab:model_complexity}.
Specifically, there are $2 H^2 + 2 N_d H + H$ additional parameters corresponding to the dimensionality of $\b{W}_h$, $\b{W}_{ze}$, $\b{W}_{zl}$, and $\b{b}$ in \cref{eq:g_detail}.
Thus, with this choice of $g$ and given $H$, the number of additional parameters of LiteralE grows in $O(N_d)$, that is, linear to the number of data relations in the KG. Furthermore, the additional space complexity of LiteralE is in $O(N_e N_d)$ as one needs to store the matrix $\mathbf{L}$. Lastly, the additional computational complexity of LiteralE is only attributed to the cost of three matrix multiplication and one vector addition.

In summary, 
with our method LiteralE, 
we propose to replace the score function $f_X(\mathbf{e}_i, \mathbf{e}_j, \mathbf{r}_k)$ from the host method $X$ with the function composition 
$$
    f_X(g(\mathbf{e}_i, \mathbf{l}_i), g(\mathbf{e}_j, \mathbf{l}_j), \mathbf{r}_k)
$$ 
as illustrated in Figure~\ref{fig:literalE}. 
This new scoring function can be trained by gradient descent 
based optimization using the same training procedure as before.

\section{Related Work}
\label{sec:related}

In the last years, several efforts to incorporate literals into KG embedding methods have been made. \cite{toutanova2015representing} and \cite{tu2017cane} make use of textual literals of entities in addition to relational embeddings. More specifically they learn additional entity embeddings from their textual description and use them in an additive term in the score function of latent distant methods. \cite{xie2016representation} and \cite{xu2016knowledge} also proposed methods to incorporate textual literals into 
latent distance methods
such as TransE by encoding textual literals with recurrent or convolutional neural networks.
\cite{xie2016image} use image literals in their model by projecting entities' image features into an entity embeddings space. However, all of those approaches do not consider numerical literals. 
MultiModal~\cite{pezeshkpour2017embedding} 
extends DistMult to also
predict the likeliness of (subject, relation, literal)-triples, by replacing the object embedding in standard DistMult by its literal embedding (where literals of different modalities are taken into account). By doing so literals are incorporated into entity embeddings 
in an implicit manner. \cite{sun2017cross}, proposes to employ literals to refine the joint embeddings in entity alignment tasks: They use literals to cluster together entities which have high literal correlations, thus only indirectly use the literal information in the entity embeddings.
In contrast to all the aforementioned works, LiteralE combines the literals into the entity embedding directly and explicitly by the function $g$ defined above.

KBLRN~\cite{garcia2017kblrn} handles literals in a separate function added to the vanilla scoring function and thus
does not incorporate literals in to the entity embeddings themselves.
The construction of features from the numerical literals
is based on a prior knowledge: the difference between 
the
numerical literals of the subject and object entity is a good predictor for a given relation.

These features then serve as input to a fixed radial basis function (RBF), which is added to the score function of the base method (DistMult).
In contrast, LiteralE incorporates literal information directly into the entity embeddings\footnote{Note, that incorporating the literal information into the embeddings also seems advantageous for entity disambiguation or clustering.}, and does not use any prior knowledge about the meaning of numerical literals.

MTKGNN~\cite{tay2017multi} extends ERMLP~\cite{dong2014knowledge} and incorporates numerical literals by introducing an additional learning task, more precisely, the task of predicting the literal value for a given entity. This multi-task learning approach of MTKGNN requires an additional \emph{attribute-specific} training procedure. 
Therefore, adding another type or modality of literals is not straightforward and costly as another learning task needs to be devised. 
Similar to MTKGNN, TransEA \cite{wu2018knowledge} extends TransE by adding a numerical attribute prediction loss to the relational loss.

Lastly, the model recently proposed by \cite{thoma2017towards} can be seen as a special case of LiteralE where the function used instead of the function $g$ defined above to combine literals of entities with their entity embeddings
is a concatenation followed by singular value decomposition. Thus, they use  a fixed function to combine the representations, whereas LiteralE employs an adaptable function and is therefore more flexible. Furthermore, they only consider image and text literals 
but no numerical literals.

\section{Experiments}
\label{sec:experiment}

In the following we will describe the training approach, the datasets, the experimental setup, and the evaluation metrics applied in our experiments.

\begin{table}[t]
  \caption{Model complexity in terms of number of parameters of methods for incorporating literals. 
  We denote the number of parameters of base models (e.g.~DistMult) with $\Gamma$. Furthermore, $Z$ is the number of hidden units in a neural network (e.g.~in LiteralE-MLP and MTKGNN's Attribute Networks). We leave out bias parameters for clarity.}
  \label{tab:model_complexity}
  
  \centering
  \begin{tabular}{ ll }
    \toprule
    \textbf{Model} &\textbf{Number of Parameters} \\
    \cmidrule{1-2}
    KBLN & $\Gamma + N_r N_d$ \\
    MTKGNN & $\Gamma + N_d H + 2(2HZ+Z)$ \\
    \midrule
    LiteralE & $\Gamma + 2H^2 + 2N_dH + H$ \\
	\bottomrule
  \end{tabular}
\end{table}

\subsection{Training}
We use the same training approach as \cite{dettmers2018conve} for all the tested methods.
That is, for every given triple $(e_i,e_j,r_k)$ in the KG, we compute the score for $(e_i,e_j^{\prime},r_k), \forall e_j^{\prime} \in \mathcal{E}$ using the 
(original or LiteralE-extended) 
scoring function $f$,
and apply the sigmoid function 
to the resulting score (i.e.~$p = \sigma \circ f$), such that it can be interpreted as probability of existence of a given triple. 

Let  $ \textbf{p} \in [0, 1]^{N_e}$ be the probability vector, collecting the resulting probabilities with respect to all $e_j^{\prime} \in \mathcal{E}$.
The model is then trained by minimizing the binary cross-entropy loss between the
probability vector $ \textbf{p}$ 
and the vector of ground truth labels $\textbf{y} \in \{0,1\}^{N_e}$ 
indicating the existence of 
triples $(e_i,e_j^{\prime},r_k), \forall e_j^{\prime} \in \mathcal{E}$ in the KG. That is, we minimize
\begin{equation}
\label{loss}
L(\textbf{p}, \textbf{y}) = -\frac{1}{N_e} \sum_{x=1}^{N_e} ({y}_x \log({p}_x) + (1-{y}_x)\log(1-{p}_x)) \enspace,
\end{equation}
where ${p}_x$ and ${y}_x$ are the predicted probability and the given truth value for the $x$-th element of our candidate set $\{(e_i,e_j^{\prime},r_k), 
e_j^{\prime} \in \mathcal{E}\}$.
We use Adam \cite{kingma2014adam} to optimize this loss function.

Note, the above procedure of
considering  all triples $(e_i,e_j^{\prime},r_k)$, $\forall e_j^{\prime} \in \mathcal{E}$ if there is any triple $(e_i,e_j,r_k)$ with head $e_i$ and relation $r_k$ in the training set
is referred to as 1-N scoring~\cite{dettmers2018conve} as for each triple, we compute scores of $N := N_e = \vert \mathcal{E} \vert$ triples. 
This is in contrast with 1-1 scoring, where
one primarily considers the training example $(e_i,e_j,r_k)$
and applies some other strategy for negative sampling (i.e.~for the generation of non-existing triples).
We refer the reader to~\cite{dettmers2018conve} for a further discussion regarding this.

\subsection{Datasets}

\begin{table}[t]
  \caption{
  Number of entities, relations, literals, and triples, for all datasets used in this paper.}
  \label{tab:dataset_stats}
  
  \centering
  \begin{tabular}{ lrrr }
    \toprule
    \textbf{Dataset} & \textbf{\textsf{FB15k}} & \textbf{\textsf{FB15k-237}} & \textbf{\textsf{YAGO3-10}} \\
    \cmidrule{1-4}
    \# Entities ($N_e$) & 14,951 & 14,541 & 123,182 \\
    \# Relations ($N_r$) & 1,345 & 237 & 37 \\
    \# Data rel.~($N_d$) & 121 & 121 & 5 \\
    \# Literals ($|\mathcal{L}|$) & 18,741 & 18,741 & 111,406 \\
    \# Relational triples & 592,213 & 310,116 & 1,089,040 \\
    \# Literal triples & 70,257 & 70,257 & 111,406 \\
\bottomrule
  \end{tabular}
\end{table}

We use three widely used datasets for evaluating link prediction performance: \textsf{FB15k}, \textsf{FB15k-237}, and \textsf{YAGO3-10}. \textsf{FB15k}~\cite{bordes2013translating} is a subset of Freebase where most triples are related to movies and sports. As discussed by \cite{dettmers2018conve}, \textsf{FB15k} has a large number of test triples which can simply be obtained by inverting training triples. This results in a biased test set, for which a simple model which is symmetric with respect to object and subject entity is capable of achieving excellent results. To address this problem, 
\textsf{FB15k-237}~\cite{toutanova2015observed} was created  by removing inverse relations from \textsf{FB15k}. \textsf{YAGO3-10}~\cite{mahdisoltani2014yago3} is a subset of the YAGO3 knowledge graph which mostly consists of triples related to people. 

In this work, we only consider numerical literals, e.g. longitude, latitude, population, age, date of birth (in UNIX time format), etc. To enrich \textsf{FB15k} and \textsf{FB15k-237} with these literals, we created a SPARQL endpoint for Freebase and extracted literals of all entities contained in \textsf{FB15k}. We further filtered the extracted literals based on their frequency, i.e~we only consider data relations $d \in \mathcal{D}$ that occur at least in 5 triples in \textsf{FB15k}. We also remove all key and ID relations
since their values are not meaningful as quantities. For \textsf{YAGO3-10}, we use numerical literals provided by YAGO3-10-plus~\cite{pezeshkpour2017embedding}, which is publicly available.\footnote{https://github.com/pouyapez/multim-kb-embeddings}
In case an entity has multiple literal values for a particular data relation, we arbitrarily select one of them. Some statistics of the datasets are provided in Table~\ref{tab:dataset_stats}.

\subsection{Experimental Setup}

\begin{table*}
  \caption{Link prediction results on \textsf{FB15k}, \textsf{FB15k-237}, and \textsf{YAGO3-10}. 
  The best values comparing our implementation of base models, KBLN, MTKGNN and LiteralE are highlighted in bold text. Only numerical literals are used in the experiments.
  }
  \label{tab:result_linkpred}

  \centering
  \begin{tabular}{ cccccc }
    \toprule
    \multicolumn{6}{c}{\textbf{\textsf{FB15k}}} \\
    \midrule
    \textbf{Models} & \textbf{MR} & \textbf{MRR} & \textbf{Hits@1} & \textbf{Hits@3} & \textbf{Hits@10} \\ 
    \midrule
    DistMult & 108 & 0.671 & 0.589 & 0.723 & 0.818 \\ 
    ComplEx & 127 & 0.695 & 0.618 & 0.744 & 0.833 \\ 
    ConvE & 49 & 0.692 & 0.596 & 0.760 & 0.853 \\ 
    \midrule
    KBLN~\cite{garcia2017kblrn} & 129 & 0.739 & 0.668 & \textbf{0.788} & 0.859 \\ 
    MTKGNN~\cite{tay2017multi} & 87 & 0.669 & 0.586 & 0.722 & 0.82 \\ 
    \midrule
    DistMult-LiteralE & 68 & 0.676 & 0.589 & 0.733 & 0.825 \\ 
    ComplEx-LiteralE & 80 & \textbf{0.746} & \textbf{0.686} & 0.782 & 0.853 \\ 
    ConvE-LiteralE & \textbf{43} & 0.733 & 0.656 & 0.785 & \textbf{0.863} \\ 
    \bottomrule
  \end{tabular}
  
  \vspace{15pt}
  
  \begin{tabular}{ cccccc }
    \toprule
    \multicolumn{6}{c}{\textbf{\textsf{FB15k-237}}} \\
    \midrule
    \textbf{Models} & \textbf{MR} & \textbf{MRR} & \textbf{Hits@1} & \textbf{Hits@3} & \textbf{Hits@10} \\ 
    \midrule
    DistMult & 633 & 0.282 & 0.203 & 0.309 & 0.438 \\ 
    ComplEx & 652 & 0.290 & 0.212 & 0.317 & 0.445 \\ 
    ConvE & 297 & 0.313 & 0.228 & 0.344 & 0.479 \\ 
    \midrule
    KBLN~\cite{garcia2017kblrn} & 358 & 0.301 & 0.215 & 0.333 & 0.468 \\ 
    MTKGNN~\cite{tay2017multi} & 532 & 0.285 & 0.204 & 0.312 & 0.445 \\ 
    \midrule
    DistMult-LiteralE & 280 & \textbf{0.317} & \textbf{0.232} & \textbf{0.348} & \textbf{0.483} \\ 
    ComplEx-LiteralE & 357 & 0.305 & 0.222 & 0.336 & 0.466 \\ 
    ConvE-LiteralE & \textbf{255} & 0.303 & 0.219 & 0.33 & 0.471 \\ 
    \bottomrule
  \end{tabular}
  
  \vspace{15pt}
  
  \begin{tabular}{ cccccc }
    \toprule
    \multicolumn{6}{c}{\textbf{\textsf{YAGO3-10}}} \\
    \midrule
    \textbf{Models} & \textbf{MR} & \textbf{MRR} & \textbf{Hits@1} & \textbf{Hits@3} & \textbf{Hits@10} \\ 
    \midrule
    DistMult & 2943 & 0.466 & 0.377 & 0.514 & 0.653 \\ 
    ComplEx & 3768 & 0.493 & 0.411 & 0.536 & 0.649 \\ 
    ConvE & 2141 & 0.505 & 0.422 & 0.554 & 0.660 \\ 
    \midrule
    KBLN~\cite{garcia2017kblrn} & 2666 & 0.487 & 0.405 & 0.531 & 0.642 \\ 
    MTKGNN~\cite{tay2017multi} & 2970 & 0.481 & 0.398 & 0.527 & 0.634 \\ 
    \midrule
    DistMult-LiteralE & 1642 & 0.479 & 0.4 & 0.525 & 0.627 \\ 
    ComplEx-LiteralE & 2508 & 0.485 & 0.412 & 0.527 & 0.618 \\ 
    ConvE-LiteralE & \textbf{1037} & \textbf{0.525} & \textbf{0.448} & \textbf{0.572} & \textbf{0.659} \\ 
    \bottomrule
  \end{tabular}
\end{table*}

We implemented LieralE
on top of ConvE's codebase, which is publicly available\footnote{https://github.com/TimDettmers/ConvE}.
The hyperparameters used in all of our experiments across all datasets are: learning rate 0.001, batch size 128, embedding size 200,
embedding dropout probability 0.2, 
and label smoothing 0.1. 
Additionally for ConvE, we used feature map dropout with probability 0.2 and projection layer dropout  with probability 0.3.
Note, that these hyperparameter values are the same as in the experiments of \cite{dettmers2018conve}.

Except for experiments with ConvE, we run all of our experiments for a maximum of 100 epochs as we observed that this is sufficient for convergence.
For ConvE, we used at most 1000 epochs, as described in the original paper~\cite{dettmers2018conve}. We apply early stopping in all of the experiments by monitoring the Mean Reciprocal Rank (MRR) metric on the validation set every three epochs.

To validate our approach and to eliminate the effect of different environment setups, we re-implemented the related models, KBLN~\cite{garcia2017kblrn}, and MTKGNN~\cite{tay2017multi} as baselines. Note that we did not re-implement KBLRN~\cite{garcia2017kblrn} 
since the sub-model
KBLN (i.e.~the KBLRN model without making use of the relational information provided by graph feature methods)
is directly comparable to LiteralE.\footnote{Note, that LiteralE could also be extended to incorporate graph features
as an additional input to $g$.
} As in \cite{dettmers2018conve}, we use a 1-N training approach, while KBLN and MTKGNN uses a 1-1 approach. Therefore, the RelNet in MTKGNN which is a neural network
is infeasible to be implemented in our environment. Thus, as opposed to neural network, we use DistMult as base model in our re-implementation of an MTKGNN-like method. While this change does not allow to evaluate the performance of the original MTKGNN model, it makes our MTKGNN-like method directly comparable to the other methods that we consider in our experiments, since it uses the same base score function.
All in all, due to these differences in the loss function and the overall framework which are necessary to make KBLN and MTKGNN comparable to LiteralE, the results we report for them 
might 
differ from those reported in the respective original papers. In addition,  we obtain slightly different results compared to~\cite{dettmers2018conve} for DistMult, ComplEx and ConvE  for all three datasets (our results are mostly comparable or slightly better and in some case worse). 
This could be attributed to the hyperparameter tuning performed in~\cite{dettmers2018conve}.

\subsection{Evaluation}

For the evaluation of the performance of the different methods on the link prediction task, we follow the standard setup used in other studies.
For each triple $(e_i,e_j,r_k)$ 
in the test set, we generate a set of corrupted triples 
by either replacing the subject entity
 $e_i$ or the object entity $e_j$ with any other entity $e' \in \mathcal{E}$.
We further compute the scores of these corrupted triples along with the score of the true triple. To evaluate the model, we rank all triples with respect to their scores and use the following standard evaluation metrics: 
Mean Rank (MR), Mean Reciprocal Rank (MRR), Hits@1, Hits@3, and Hits@10.

\section{Results}
\label{sec:results}

\subsection{Link Prediction}

The results of our experiments for link prediction are summarized in Table~\ref{tab:result_linkpred}. In general, LiteralE improves the base models (DistMult, ComplEx, and ConvE) significantly. For instance, we found that implementing LiteralE on top of DistMult improves the MRR score by 
0.74\%, 12.41\%, and 2.7\% for the \textsf{FB15k},
\textsf{FB15k-237}, and \textsf{YAGO3-10} dataset, respectively. We also observed that the improvements brought by LiteralE 
when combined with
ComplEx and ConvE are not as impressive as for DistMult, which might be attributed to the fact that these base models already achieve higher performance than DistMult. Compared to other methods that incorporate literals, namely KBLN and MTKGNN, LiteralE achieves a competitive or even better performance in our experiments. Moreover, note that, LiteralE directly and explicitly modifies the embedding vectors, whereas KBLN and MTKGNN do not. Thus, LiteralE embeddings could be more useful for tasks other than link prediction. 
This will be discussed further in \Cref{subsec:nearest_neighbor}.

\subsection{Comparison to a Simple LiteralE Baseline}
\label{subsec:comparison_g_linear}

\begin{table}[t]
  \caption{The link prediction performance 
  of LiteralE 
  employing a simple linear transformation 
  $g_\text{lin}$.}
  \label{tab:ablation}
  \centering
  \begin{tabular}{ ccccccc }
    \toprule
    \textbf{Datasets} & \textbf{Functions} & \textbf{MRR} & \textbf{Hits@1} & \textbf{Hits@10} \\
    \midrule
    \textsf{FB15k} 
    & DistMult-$g_\text{lin}$ & 0.583 & 0.476 & 0.771 \\
    & ComplEx-$g_\text{lin}$ & \textbf{0.765} & \textbf{0.705} & \textbf{0.871} \\
    & ConvE-$g_\text{lin}$ & 0.66 & 0.556 & 0.836 \\
    \midrule
    \textsf{FB15k-237} 
    & DistMult-$g_\text{lin}$ & \textbf{0.314} & \textbf{0.228} & \textbf{0.483} \\
    & ComplEx-$g_\text{lin}$ & 0.299 & 0.214 & 0.467 \\
    & ConvE-$g_\text{lin}$ & \textbf{0.314} & \textbf{0.228} & \textbf{0.483} \\
    \midrule
    \textsf{YAGO3-10} 
    & DistMult-$g_\text{lin}$ & 0.504 & 0.422 & 0.653 \\
    & ComplEx-$g_\text{lin}$ & \textbf{0.509} & \textbf{0.433} & 0.653 \\
    & ConvE-$g_\text{lin}$ & 0.506 & 0.422 & \textbf{0.664} \\    
    \bottomrule
  \end{tabular}
\end{table}

To validate our choice of the function $g$, we compare the performance of LiteralE with the $g$ proposed in \Cref{sec:literale} to its variant based on a simple (but still learnable) linear transformation, dubbed $g_\text{lin}$. That is, $g_\text{lin}: \R^{H} \times \R^{N_d} \to \R^{H}$ is defined by $\b{e}, \b{l} \mapsto \b{W}^\text{T} [\b{e}, \b{l}]$, where $\b{W} \in \R^{H+N_d \times H}$ is a learnable weight matrix. The results are presented in \Cref{tab:ablation} (cf. \Cref{tab:result_linkpred}).

The proposed $g$ leads to better results than $g_\text{lin}$ in 5 out of 9 experiments. While LiteralE with $g$ provides a consistent performance improvement for all base models, DistMult-$g_{lin}$ shows a decreased  performance compared to DistMult on \textsf{FB15k}. 
 This might be explained by the fact that -- as \cite{toutanova2015observed} already reported -- \textsf{FB15k} contains triples in the test set that have an inverse analog (i.e.~the triple resulting from  changing the position of  subject and object entity) in the training set. The prediction for such triples can get difficult if the inverse has a different label. Since the vanilla DistMult already has difficulties in modeling asymmetric relations  on \textsf{FB15k}, adding literals using a naive $g_\text{lin}$  might  only introduce noise, resulting in even lower performance.
On the other hand,  $g_\text{lin}$  leads to better results than $g$ in combination with ComplEx on \textsf{FB15k}.

In general, the results show that for performance-maximization purpose, it makes sense to investigate the performance of LiteralE in combination with different transformation functions. Given the right choice of transformation function for incorporating literals, LiteralE always improves the performance of the base model. 

\subsection{Experiment with Text Literals}
\label{subsec:text_experiment}

\begin{table}[t]
  \caption{Link prediction results for DistMult-LiteralE on \textsf{FB15k-237}, with both numerical and text literals. ``DM'' and ``L'' stand for DistMult and LiteralE respectively, while ``N'' and ``T'' denote the usage of numerical and text literals, respectively.}
  \label{tab:result_text_lit}
  \centering
  \begin{tabular}{ ccccccc }
    \toprule
    \textbf{Models} & \textbf{MRR} & \textbf{Hits@1} & \textbf{MRR Incr.} \\
    \midrule 
    DM & 0.241 & 0.155 & - \\
    DM-L (N) & 0.317 & 0.232 & 0+31.54\% \\
    DM-L (N+T) & \textbf{0.32} & \textbf{0.234} & \textbf{+32.78\%} \\
    \bottomrule
  \end{tabular}
\end{table}

\begin{table*}[ht!]
  \caption{Comparison of nearest neighbors of selected entities from \textsf{FB15k-237} embedded in (i) DistMult's latent space, (ii) KBLN's latent space, (iii) MTKGNN's latent space, (iv) the literal space, where each entity is represented only by its literals, and (v) the DisMult-LiteralE's latent space.}
  \label{tab:nearest_neighbor}
  \centering
  \begin{tabularx}{\linewidth}{ l l X }
    \toprule
    \textbf{Entity} & \textbf{Methods} & \textbf{Nearest Neighbors} \\
    \midrule
    North America & DistMult & Latin America, Pyrenees, Americas \\
    \cmidrule{2-3}
    & KBLN & House of Hanover, House of Stuart, House of Romanov \\
    \cmidrule{2-3}
    & MTKGNN & Latin America, Panama City, Pyrenees \\
    \cmidrule{2-3}
    & Num. lits. only & Soviet Union, Latin America, Africa \\
    \cmidrule{2-3}
    & LiteralE & \textbf{Americas}, \textbf{Latin America}, \textbf{Asia} \\
    \midrule
    Philippines & DistMult & Peru, Thailand, Kuwait \\
    \cmidrule{2-3}
    & KBLN & House of Romanov, House of Hanover, House of Stuart \\
    \cmidrule{2-3}
    & MTKGNN & Thailand, Kuwait, Peru \\
    \cmidrule{2-3}
    & Num. lits. only & Peru, Poland, Pakistan \\
    \cmidrule{2-3}
    & LiteralE & \textbf{Thailand}, \textbf{Taiwan}, \textbf{Greece} \\
    \midrule
    Roman Republic & DistMult & Republic of Venice, Israel Defense Force, Byzantine Empire \\
    \cmidrule{2-3}
    & KBLN & Republic of Venice, Carthage, Retinol \\
    \cmidrule{2-3}
    & MTKGNN & Republic of Venice, Carthage, North Island \\
    \cmidrule{2-3}
    & Num. lits. only & Alexandria, Yerevan, Cologne \\
    \cmidrule{2-3}
    & LiteralE & \textbf{Roman Empire}, \textbf{Kingdom of Greece}, \textbf{Byzantine Empire} \\
    \bottomrule
  \end{tabularx}
\end{table*}

LiteralE, as  described in \Cref{sec:literale} can easily be extended to other types of literals, e.g.~text and images. In this section 
this is briefly demonstrated for  text literals.
First, let us assume that text literals are represented by vectors in $\R^{N_t}$, i.e.~as resulting from document embedding techniques \cite{le2014distributed}.\footnote{We use spaCy's pretrained GloVe embedding model. Available at \url{https://spacy.io}} Subsequently, let us redefine $g$ to be a function  mapping $\R^H \times \R^{N_d} \times \R^{N_t}$ to $\R^H$. Specifically, we redefine $\b{W}_h$ (\cref{eq:g_detail}) to be in $\R^{H+N_d+N_t \times H}$ and employ an additional gating weight matrix $\b{W}_{zt} \in \R^{N_t \times H}$ to handle the additional text literal. Note, that this simple extension scheme can be used to extend LiteralE to incorporate literals of any other type (e.g.~image literals) as long as those literals are encoded as vectors in $\R^N$, for some $N$.

The results for extending DistMult-LiteralE with the entities' text literals (i.e.~the entity description) are presented in \Cref{tab:result_text_lit}. We found that incorporating text literals results in a further increase of the link prediction performance of DistMult on \textsf{FB15k-237}.

\subsection{Nearest Neighbor Analysis}
\label{subsec:nearest_neighbor}

For a further qualitative investigation, we present 
the nearest neighbors of some entities in the space of literals, the latent space learned by (i) DistMult, (ii) KBLN, (iii) MTKGNN, and (iv) DisMult-LiteralE in Table~\ref{tab:nearest_neighbor}.\footnote{The base model for all of these methods is DistMult.}

In the embedding space of DistMult, geographical entities such as \texttt{North America} and \texttt{Philippines} are close to other entities of the same type. However, these neighboring entities are not intuitive, e.g.~\texttt{North America} is close to \texttt{Pyrenees}, whereas \texttt{Philippines} is close to \texttt{Peru} and \texttt{Kuwait}. When we inspected the embedding space of DistMult-LiteralE that also takes literals information into account, these nearest neighbors (shown in bold font in Table~\ref{tab:nearest_neighbor}) become more intuitive, i.e~they consist of entities geographically close to each others.
Furthermore, we found that DistMult-LiteralE's embeddings show clear qualitative advantage compared to that of vanilla DistMult also for entities from other types, e.g. comparing the nearest neighbors of \texttt{Roman Republic} which is of type `empire'.
In contrast, KBLN's embeddings tend to be close to the embeddings of unrelated entities: both \texttt{North America} and \texttt{Philippines} are close to the entities \texttt{House of Romanov}, \texttt{House of Hanover}, and \texttt{House of Stuart}, while \texttt{Roman Republic} is close to \texttt{Retinol}. 
Similarly, the embeddings of MTKGNN are also close to the embedding of unrelated entities, e.g., \texttt{North America} is close to \texttt{Pyrenees} and \texttt{Roman Repulic} is close to \texttt{North Island}.
This findings demonstrates the advantage of incorporating literals 
on the embedding level (as done by LiterelE)
over incorporating them at the score or loss function (as done by KBLN and MTKGNN, respectively).

When inspecting the nearest neighborhood of the same entities when represented only by their literal vectors,
it becomes clear that these
vectors themselves are already containing useful information indicating the closeness of similar entities. For example, geographical entities have \texttt{longitude} and \texttt{latitude} literals, while city, nation, and empire entities have \texttt{date\_founded} and \texttt{date\_dissolved} literals,
which can explain the closeness of two entities given only their literal vectors. Note however, that
the nearest neighbours in the literal space do not coincide with and are less informative than the nearest neighbours in the LiteralE embedding space.

All in all, our observations suggest that integrating the literal information into entity embeddings 
indeed improves their quality, which 
makes LiteralE embeddings
promising 
for 
entity resolution and clustering tasks.  

\section{Conclusion and Future Work}
\label{sec:conclusion}

In this paper, we introduced LiteralE: a simple method to incorporate literals into 
latent feature methods for knowledge graph analysis. 
It corresponds to a learnable function that merges entity embeddings with their literal information available in the knowledge graph. The resulting literal-enriched latent features can replace the vanilla entity embedding in any latent feature method, without any further modification. Therefore, 
LiteralE can be seen as an universal extension module. 
We showed that
augmenting various state-of-the-art  models (DistMult, ComplEx, and ConvE) with LiteralE significantly improves their link prediction performance.
Moreover, as exemplarily  demonstrated for text literals, LiteralE can be easily extended other types of literals. 
In future work, LiteralE shall be further be extended to accommodate literals from
the image domain.
This can be achieved by extracting 
latent representations
from images (for example with convolutional neural networks),
and providing them as additional inputs to LiteralE for merging them with the vanilla entitiy embeddings.
Furthermore, 
our finding that LiteralE improves the quality of the entity embeddings makes it a promising candidate for improving other tasks in the field of knowledge graph analysis,
such as entity resolution and knowledge graph clustering.

\bibliographystyle{named}
\small
\bibliography{kge_literals}

\end{document}